\documentclass{isprs} 
\usepackage{subfigure}
\usepackage{setspace}
\usepackage{geometry} 
\usepackage{epstopdf}
\usepackage[labelsep=period]{caption}  
\usepackage[british]{babel} 
\usepackage[hang]{footmisc}

\begin{document}

\title{Synthetic Dataset Generation for Partially Observed Indoor Objects
}

\author{
 Jelle Vermandere\textsuperscript{1}, 
 Maarten Bassier\textsuperscript{1},
 Maarten Vergauwen\textsuperscript{1}}

\address{
	\textsuperscript{1} KU Leuven, Department of Civil Engineering, Ghent, Belgium \\
    (jelle.vermandere, maarten.bassier, maarten.vergauwen)@kuleuven.be\\
}

\abstract{
Learning-based methods for 3D scene reconstruction and object completion require large datasets containing partial scans paired with complete ground-truth geometry. However, acquiring such datasets using real-world scanning systems is costly and time-consuming, particularly when accurate ground truth for occluded regions is required. 

In this work, we present a virtual scanning framework implemented in Unity for generating realistic synthetic 3D scan datasets. The proposed system simulates the behaviour of real-world scanners using configurable parameters such as scan resolution, measurement range, and distance-dependent noise. Instead of directly sampling mesh surfaces, the framework performs ray-based scanning from virtual viewpoints, enabling realistic modelling of sensor visibility and occlusion effects. In addition, panoramic images captured at the scanner location are used to assign colours to the resulting point clouds.

To support scalable dataset creation, the scanner is integrated with a procedural indoor scene generation pipeline that automatically produces diverse room layouts and furniture arrangements. Using this system, we introduce the \textit{V-Scan} dataset, which contains synthetic indoor scans together with object-level partial point clouds, voxel-based occlusion grids, and complete ground-truth geometry. The resulting dataset provides valuable supervision for training and evaluating learning-based methods for scene reconstruction and object completion.
}

\maketitle

\section{Introduction}

Predicting missing geometry in partially scanned 3D environments is essential for many applications in the architecture, engineering and construction (AEC) industry. Accurate reconstruction of occluded regions supports tasks such as digital twin generation, progress monitoring, and structural analysis. In recent years, deep learning based scene- and object-completion networks have made significant progress in recovering missing geometry from partial observations \cite{vermandere_geometry_2025}. However, training such models requires large amounts of high-quality data containing both partial scans and corresponding ground-truth geometry.

Acquiring such datasets from real-world scans remains time-consuming and expensive. Capturing complete ground-truth geometry typically requires multiple scanning positions or carefully controlled acquisition environments, which limits scalability. As a result, many existing real-world datasets provide only partial observations without accurate ground-truth information for occluded regions, restricting their usefulness for supervised geometric completion tasks.

Synthetic datasets offer a promising alternative. A common approach is to directly sample points on the surfaces of 3D meshes \cite{de_geyter_automated_2022}. While efficient, this method produces unrealistic point distributions and often includes surfaces that would be invisible to a real scanner. Consequently, these datasets fail to capture important physical constraints of real scanning systems, such as visibility, range limitations, and measurement noise.

To address these challenges, we propose a virtual scanning framework implemented in Unity that simulates the behaviour of real-world scanning devices. Instead of directly sampling mesh surfaces, the system performs ray-based scanning from configurable virtual viewpoints, allowing realistic modelling of sensor characteristics such as measurement range, angular resolution, occlusion, and distance-dependent noise. In addition, panoramic imagery captured at the scanner location is used to assign colours to the scanned points, producing visually consistent point clouds that reflect the lighting conditions of the virtual environment.

To support large-scale dataset generation, the framework is integrated with a procedural scene generation system that automatically creates indoor layouts and populates them with furniture objects. Each generated scene can be scanned automatically using the virtual scanner, producing realistic partial observations of objects and environments. Furthermore, voxel-based occlusion grids are computed at both scene and object level, enabling explicit representation of visibility within the scanned environment.

Using this pipeline, we introduce the \textit{V-Scan} dataset, a synthetic dataset of indoor scans containing both partial observations and complete ground-truth geometry. Each scene includes a furnished scan, an empty scene scan providing ground truth geometry, object-level partial point clouds, and voxel-based occlusion annotations. This combination provides valuable supervision signals for learning-based reconstruction and completion methods.

The main contributions of this work are:
\begin{itemize}
    \item \textbf{Virtual Scanner for Unity:} a plug-and-play virtual scanner that simulates real-world scanning behaviour and can be integrated into arbitrary Unity scenes.
    \item \textbf{Procedural Scene Generation:} an automated pipeline for generating indoor layouts and populating them with diverse furniture configurations.
    \item \textbf{Occlusion Voxel Grids:} computation of full-scene and per-object occlusion grids that explicitly represent visibility within the scanned environment.
    \item \textbf{Synthetic Dataset:} \textit{V-Scan}, a dataset of partially scanned indoor scenes and objects paired with complete ground-truth geometry.
\end{itemize}

Overall, the proposed framework bridges the gap between simple surface sampling and costly real-world scanning by combining procedural scene generation with physically plausible virtual scan simulation.

\begin{table*}[t]
\centering
\caption{Comparison of representative 3D scanning devices ranging from survey-grade terrestrial scanners to mobile and consumer devices.}
\begin{tabular}{llrrr}
\hline
Device & Type & Range & $\delta_r$ & Scan Rate \\
\hline
Leica P30 & TLS & up to 270 m & $1.2mm + 10ppm$ & 1MHz \\
NavVis VLX & Mobile SLAM LiDAR & up to 50 m & $6mm/500m^22$ & 1.28MHz \\
Leica BLK360 & Compact terrestrial LiDAR & up to 45 m & $4mm @ 10m$ & 0.68MHz \\
iPhone Pro LiDAR & Mobile ToF LiDAR & $\sim$5 m & $\sim1-2 cm$ & N/A (depth sensing) \\
\hline
\end{tabular}
\label{tab:scanner_comparison}
\end{table*}

\begin{figure*}[h]
    \centering
    \includegraphics[width=1\linewidth]{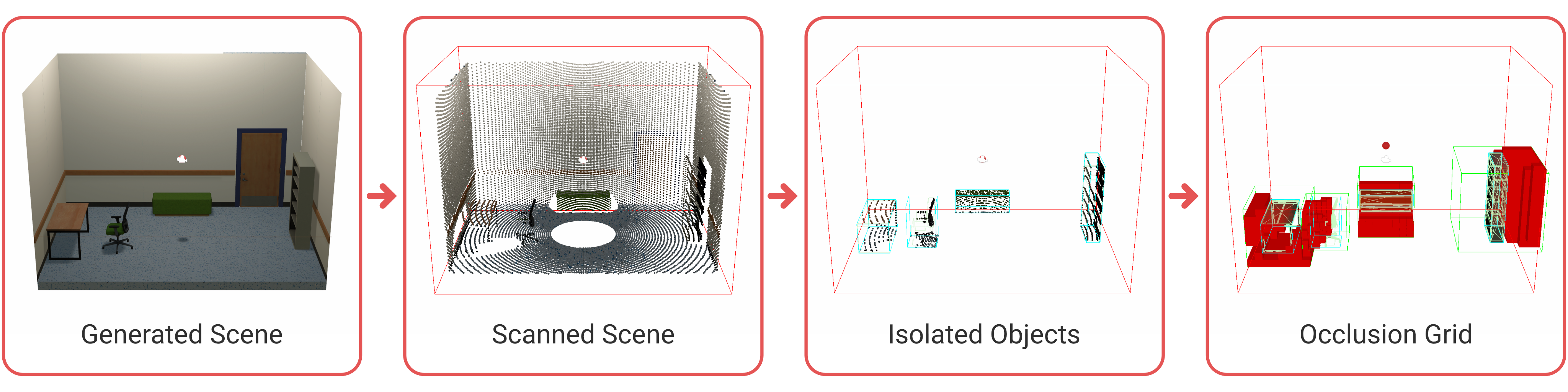}
    \caption{Virtual scanner overview with: input }
    \label{fig:method}
\end{figure*}

\section{Background and Related Work}

\subsection{Full-Scene Indoor 3D Scan Datasets}

Large-scale indoor 3D datasets have been instrumental in advancing research in scene understanding and reconstruction. Early efforts include \textit{SUN RGB-D} \cite{song_sun_2015}, which provides over 10,000 RGB-D images with 2D and 3D annotations, including room layouts and object orientations. \textit{Matterport3D} \cite{chang_matterport3d_2017} and \textit{ScanNet} \cite{dai_scannet_2017} capture detailed indoor scenes using handheld or structured RGB-D scanning pipelines, while \textit{Stanford 2D-3D-S} \cite{armeni_joint_2017} provides multi-building semantic annotations with RGB images, depth maps, and 3D meshes. More recent datasets, such as \textit{ARKitScenes} \cite{baruch_arkitscenes_2022} and \textit{ScanNet++} \cite{yeshwanth_scannet_2023}, extend these efforts with mobile LiDAR acquisition and improved scene geometry, enabling tasks such as semantic segmentation, scene reconstruction, and 3D perception.  

Despite their scale, most real-world datasets provide only observed geometry, leaving occluded regions unscanned. This limits their use for supervised learning tasks that require complete 3D ground truth, such as geometric completion or reconstruction from partial observations.

\subsection{Single-Object 3D Scans}

Complementing scene-level datasets, several works focus on high-quality 3D scans of individual objects. The \textit{Redwood Object Scan Dataset} \cite{choi_large_2016} provides over 10,000 real-world object scans captured by non-expert operators, while \textit{Pix3D} \cite{sun_pix3d_2018} pairs images with aligned 3D models for furniture objects. More recent datasets include \textit{Google Scanned Objects} \cite{downs_google_2022}, which offers high-fidelity models suitable for robotics simulation, and \textit{OmniObject3D} \cite{wu_omniobject3d_2023}, containing 6,000 real-scanned objects across 190 categories with textures, point clouds, and videos. These datasets enable research in 3D reconstruction, novel-view synthesis, and neural surface modelling.

\subsection{Synthetic 3D Scene and Object Datasets}

Synthetic datasets address limitations of real-world capture by providing complete geometry and dense annotations. Scene-level synthetic datasets include \textit{SUNCG} \cite{song_semantic_2016}, used to train SSCNet for volumetric occupancy and semantic prediction, \textit{SceneNet RGB-D} \cite{mccormac_scenenet_nodate}, and \textit{InteriorNet} \cite{li_interiornet_2018}, both offering millions of photorealistic RGB-D frames from procedurally generated scenes. The \textit{Replica} dataset \cite{straub_replica_2019} provides highly detailed, photo-realistic reconstructions suitable for embodied agent research.  

At the object level, repositories such as \textit{ShapeNet} \cite{chang_shapenet_2015}, \textit{ModelNet} \cite{wu_3d_2015}, and \textit{Amazon Berkeley Objects (ABO)} \cite{collins_abo_2022} provide extensive collections of 3D CAD models or product scans. These datasets support synthetic point cloud generation, shape analysis, single-object reconstruction, and deep learning on 3D data. Importantly, synthetic approaches allow control over visibility and occlusion, which is difficult to achieve in real-world scans.

\subsection{Real-World 3D Scanning Systems}

Real-world 3D acquisition systems vary widely in accuracy, range, and mobility. Survey-grade terrestrial laser scanners such as the \textit{Leica ScanStation P30} \cite{noauthor_leica_nodate} provide millimetre-level accuracy and long measurement ranges but require static tripod-based scanning and multi-scan registration. More portable terrestrial scanners, such as the \textit{Leica BLK360} \cite{noauthor_leica_nodate-1}, offer similar scanning principles in a compact form factor with reduced range. These static systems often suffer from large occlusions due to their limited setups. Mobile mapping systems like the \textit{NavVis VLX} \cite{noauthor_navvis_nodate} use LiDAR combined with SLAM to capture large indoor environments while the operator moves through the space, enabling faster acquisition and a higher coverage at slightly lower accuracy. At the consumer level, devices such as the \textit{iPhone Pro} \cite{abdel-majeed_indoor_2024} integrate short-range LiDAR sensors designed primarily for augmented reality and coarse 3D reconstruction. The specifications of each system are shown in table~\ref{tab:scanner_comparison}.

\subsection{Virtual Scanning and LiDAR Simulation}

Recent work has explored virtual scanning frameworks to simulate the acquisition process itself. Early efforts, such as the laser scanner simulator by \cite{blume_simulating_2006}, provide efficient, real-time range measurements for wheeled robots. \textit{Popovas et al.} \cite{popovas_virtual_2021} developed a virtual scanning system for training and educational purposes, while \textit{HELIOS++}\cite{winiwarter_virtual_2022} focusses on aerial LiDAR simulation for terrain mapping and forestry analysis. By simulating viewpoint-dependent visibility and sensor behaviour, these methods allow generation of realistic partial scans that mirror real-world scanning conditions.

While these frameworks provide accurate sensor simulation, they are typically designed for surveying or robotics applications rather than scalable dataset generation for learning-based reconstruction tasks. In particular, they generally do not provide mechanisms for isolating individual objects, generating object-level ground truth, or explicitly computing occlusion information within complex indoor scenes.

The framework proposed in this work focuses specifically on generating training datasets for indoor scene understanding and object completion tasks. By integrating a configurable virtual scanner with object isolation and occlusion analysis, the system enables the generation of partial scans paired with complete ground-truth geometry. Table \ref{tab:scanner_comparison_frameworks} shows an overview of the different virtual scanner capabilities.

\begin{table}[h!]
\centering
\caption{Comparison of virtual scanning frameworks.}
\begin{tabular}{lcc}
\hline
Framework &  Indoor Scenes & Object-level GT \\
\hline
Blume et al.  & Yes & No \\
HELIOS++  & Limited & No \\
Popovas et al.  & Yes & No \\
V-Scan (ours)  & Yes & Yes \\
\hline
\end{tabular}

\label{tab:scanner_comparison_frameworks}
\end{table}

\subsection{Occlusion Modelling and Visibility Analysis}

Explicit modelling of occlusions is crucial for reconstruction, completion, and robotic perception tasks. Occlusion reasoning methods identify regions not visible from a given viewpoint, improving navigation, mapping, and planning in cluttered environments. For instance, \cite{zhu_occlusion_2020} demonstrate how integrating occlusion awareness enhances perception and task performance. Incorporating occlusion modelling into virtual scanning frameworks enables the creation of realistic partial scans with ground-truth visibility, providing valuable supervision for learning-based reconstruction and completion methods.

\subsection{Procedural Room Generation}
To create a dataset, rather than handcrafting 3D scenes one-by-one, procedural and generative methods have been the go to standard for their ability to generate a large amount of variable data efficiently. Something which is necessary for training computer vision and 3D understanding models. \textit{Infinigen Indoors} \cite{raistrick_infinigen_nodate} extends a Blender-based procedural framework to photorealistic indoor scenes by combining a large library of furniture, architectural elements, and household objects with a constraint-based arrangement system, allowing the automated generation of diverse layouts that satisfy spatial and semantic rules. Building on the idea of controllable generation, \textit{RoomPilot} \cite{chen_roompilot_2025} introduces a multi-modal Indoor Domain-Specific Language (IDSL) that translates textual descriptions or CAD floor plans into structured scene layouts, synthesizing interactive environments with realistic object behaviours and improved physical consistency. Complementing rule-based and deterministic approaches, \textit{DiffuScene} \cite{tang_diffuscene_2024} employs a de-noising diffusion model over unordered 3D object attributes, enabling the generation of diverse and physically plausible indoor scenes while maintaining symmetry, semantic coherence, and geometric realism. Together, these works highlight the spectrum of modern indoor scene generation techniques, from constraint-driven procedural systems to learned generative models, providing both controllability and realism for downstream applications such as embodied AI, scene completion, and text-conditioned synthesis.

\section{Virtual Scanner}
\label{sec:virtual_scanner}

Our goal is to generate realistic 3D scans of virtual environments that emulate physical scanner behaviour. This is performed by first determining the scan vectors and then ray-casting against the virtual scene. The system is also able to create 3D coloured pointclouds of the whole scene, isolate all the non-static objects and calculate the per-object scene occlusion as seen in Figure \ref{fig:method}.

\subsection{Virtual Scanner Setup}

\begin{figure}[h]
    \centering
    \includegraphics[width=1\linewidth]{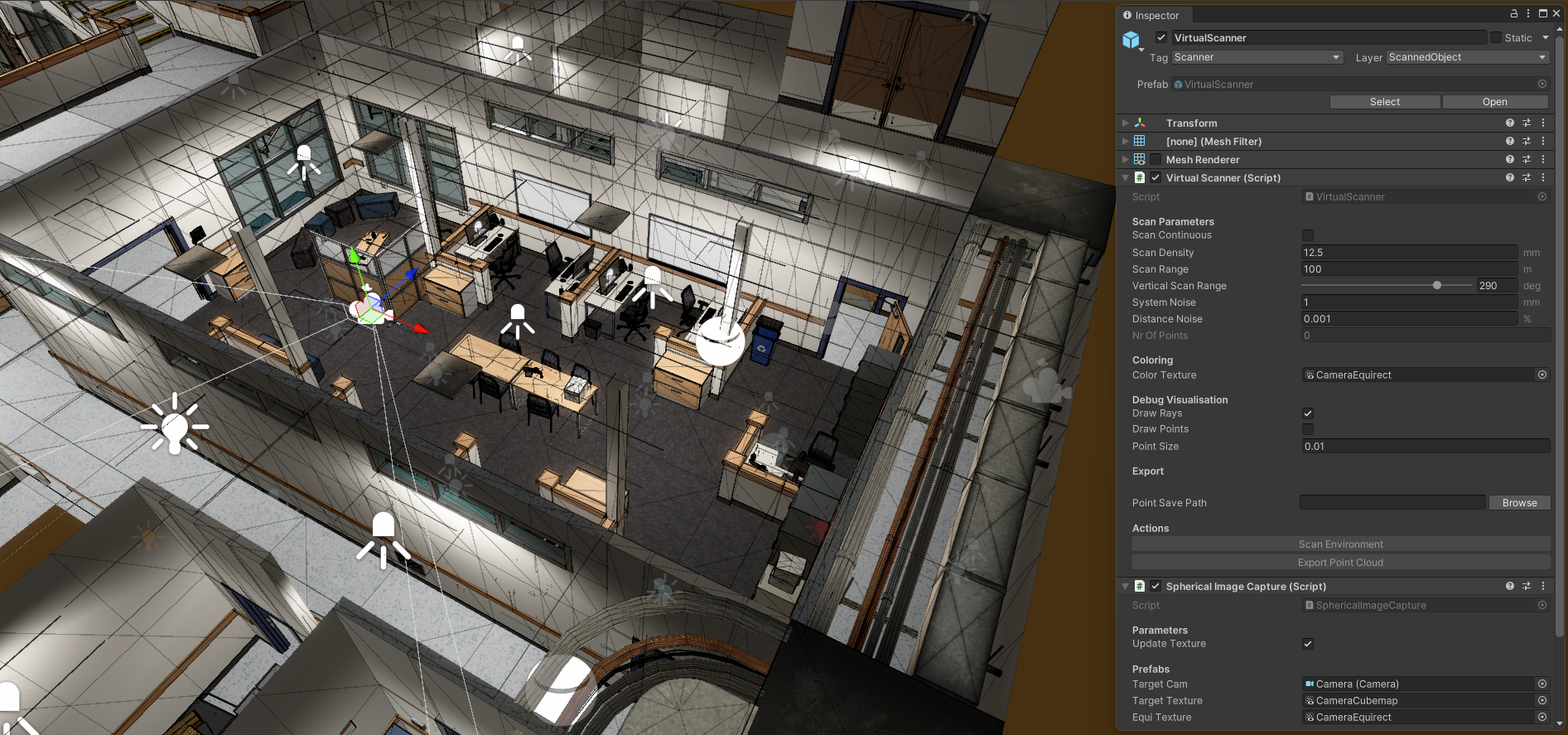}
    \caption{Virtual scanner setup in Unity with configurable parameters.}
    \label{fig:scanner_scene}
\end{figure}

The virtual scanner simulates real-world scanning devices using configurable parameters: scan density (mm per 10\,m), maximum range (m), vertical field-of-view (degrees), system error (mm), and distance-dependent error (\%). Presets emulate commercial devices such as the Leica P30, NavVis VLX, BLK360, and iPhone Pro LiDAR. These parameters determine scan resolution, coverage, and measurement precision. Figure~\ref{fig:scanner_scene} illustrates the scanner positioned in a Unity scene and the parameters on the right.

\subsection{Scan Vector Generation}

\begin{figure}
    \centering
    \includegraphics[width=1\linewidth]{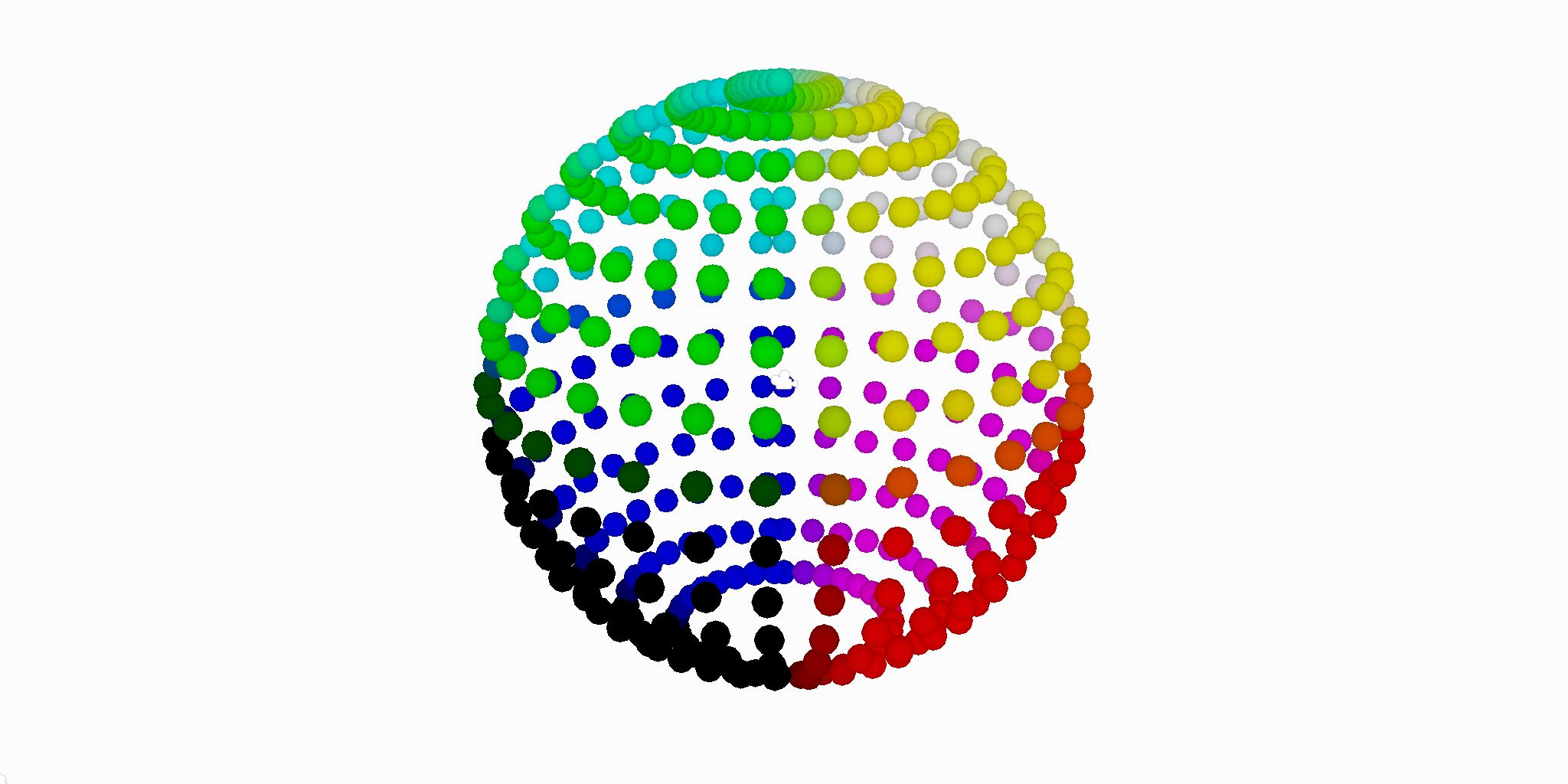}
    \caption{The scan vectors that are generated at the start of the scan, coloured according to their direction.}
    \label{fig:scanvectors}
\end{figure}

Scanning begins by generating a spherical array of vectors representing individual rays. Vectors are distributed uniformly along horizontal disks stacked vertically, starting from a minimum angle to simulate the blind zone beneath the sensor (Figure \ref{fig:scanvectors}). Equal angular spacing in both vertical and horizontal directions ensures uniform coverage and allows efficient handling of polar regions, while maintaining computational efficiency compared to sequential vertical sweeps.

\subsection{Ray-casting and Point Cloud Acquisition}

Each vector is cast from the scanner origin until it intersects a scene object. Intersection points are recorded with their 3D position and surface normal. Gaussian noise is applied to simulate system error, and a distance-dependent error is included to approximate real measurement inaccuracies. Points beyond the maximum range are discarded. The result is a partial, noise-affected point cloud representing the observed scene geometry.

\subsection{Point Colouring via Panoramic Capture}

\begin{figure}
    \centering
    \includegraphics[width=1\linewidth]{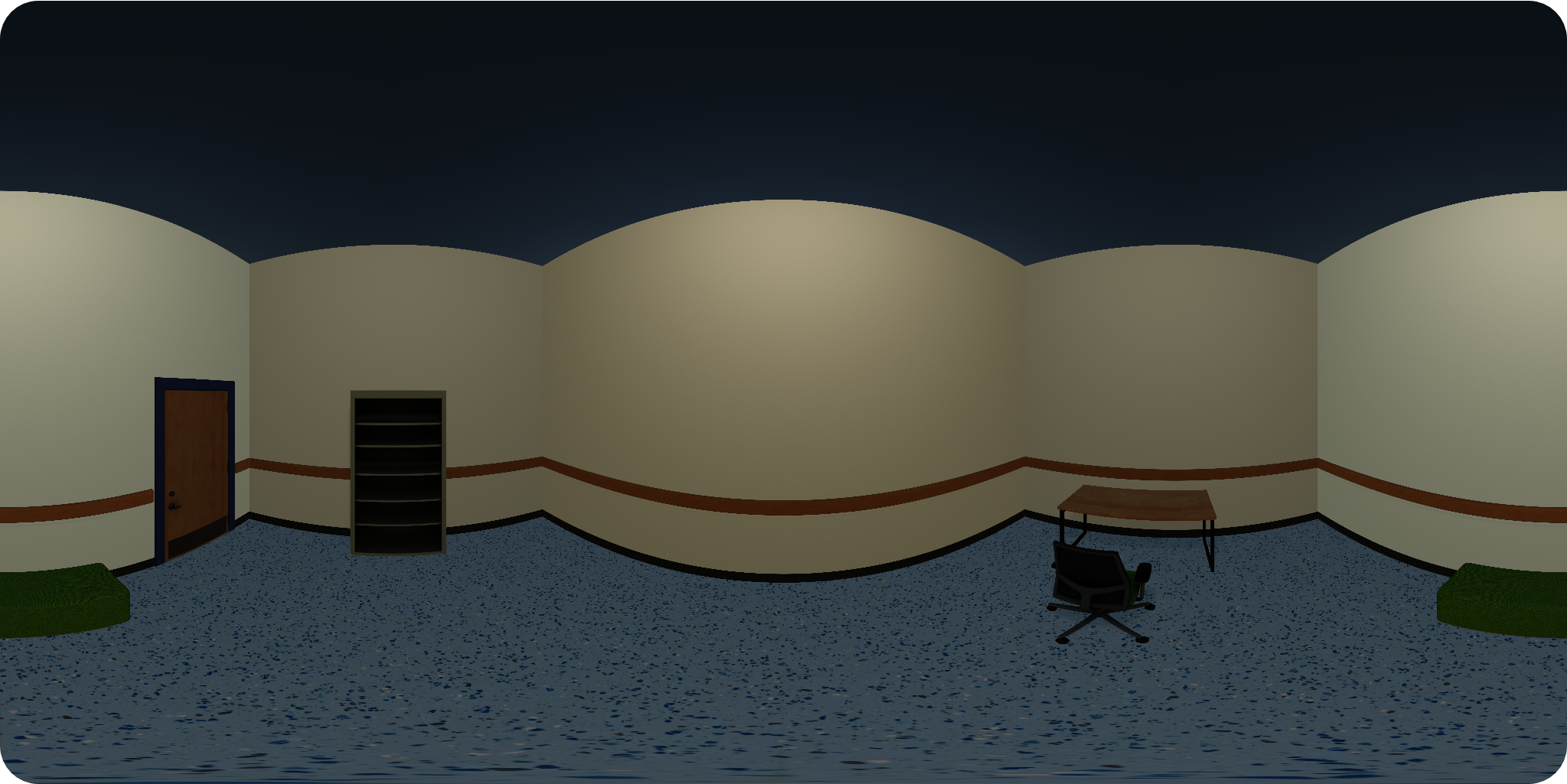}
    \caption{A equirectangular projection of the panoramic image captured by the virtual scanner}
    \label{fig:pano}
\end{figure}

To capture realistic appearance, a 360° panoramic image is generated at the scanner’s position using a cubemap converted to an equirectangular projection (Figure \ref{fig:pano}). Each scanned point is assigned a colour by mapping its angular coordinates to the corresponding pixel in the panorama. The resulting coloured point cloud preserves lighting, shading, and material information from the virtual environment, and can be exported with positions, normals, and colours for downstream tasks.

\subsection{Object-Level Scanning}

Each non-structural object in the scene is assigned a ScannedObject script, enabling isolation of points within its oriented bounding box (OBB). The OBB can be slightly expanded to ensure points near the edges are included. This facilitates exporting per-object partial scans along with bounding boxes defined by the eight 3D corner points.

\subsection{Occlusion Computation}

\begin{figure}
    \centering
    \includegraphics[width=1\linewidth]{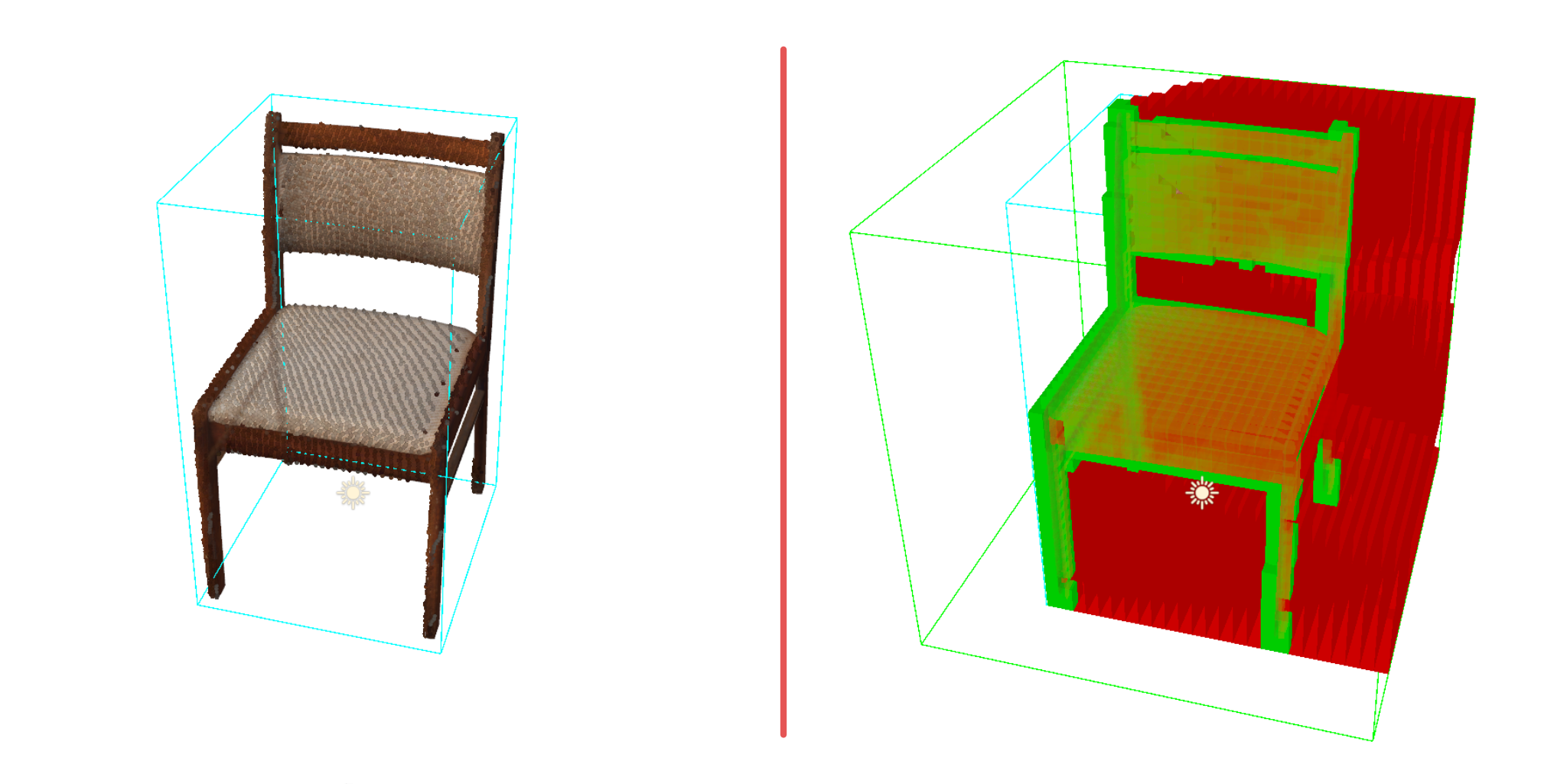}
    \caption{Partially scanned object (left) and Object-level occlusion grid showing the occupied voxels (green) and the occluded voxels(red) (right)}
    \label{fig:occlusion}
\end{figure}

Occlusion is computed both at the object and scene level using a voxel-based approach. A cubic voxel grid with a set resolution is defined by encapsulating the entire OBB, and rays are cast from the scanner origin to each voxel centre. Voxels intersected by other geometry are marked as occluded. The resulting grids are exported as sparse voxel representations (Figure \ref{fig:occlusion}), providing ground-truth visibility information for each object as well as for the entire scene.

\subsection{Empty Scene Scanning and Export}

\begin{figure}
    \centering
    \includegraphics[width=1\linewidth]{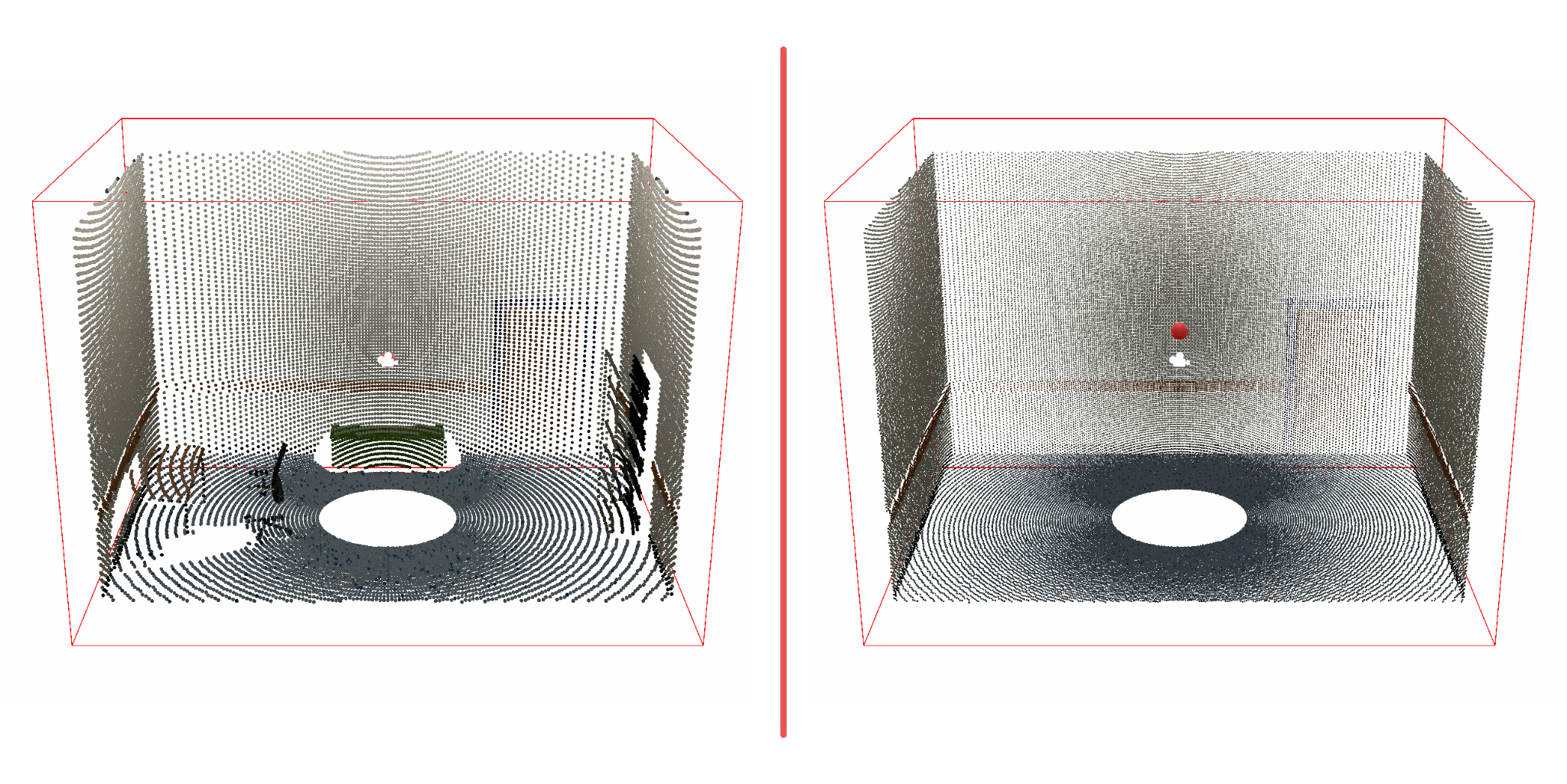}
    \caption{The scanned scene (left) and the empty rescanned scene (right)}
    \label{fig:empty}
\end{figure}

To generate complete ground-truth geometry of the empty scene, the scene is rescanned with all non-structural objects removed (Figure \ref{fig:empty}). This produces a full empty-scene point cloud and panorama, which can be exported alongside the separate object scans. The output of the pipeline includes:

\begin{itemize}
    \item Full scene scans (with and without objects), including colours and normals.
    \item Equirectangular Panoramic images (with and without objects).
    \item Full-scene voxel occlusion grids.
    \item Per-object partial point clouds, oriented bounding boxes, occlusion grids, and ground-truth meshes.
\end{itemize}

\section{Dataset}

In this section we describe the generation of the \textit{V-Scan} dataset, which consists of synthetic indoor scenes scanned using the virtual scanning framework described in the previous section. The dataset is generated fully procedurally in Unity, enabling large numbers of diverse scenes to be produced automatically. Each scene contains structural elements such as floors, walls, and ceilings, as well as non-static objects such as furniture and other technical equipment. The dataset is structured to contain ground truth, both per object and per scene.

\subsection{Procedural Layout Generation}

Indoor scenes are generated using a procedural room generation algorithm implemented in Unity. The layout generation begins by sampling the room dimensions from a predefined range, producing rectangular floor plans with widths and lengths between configurable minimum and maximum values. The room is discretized using a regular grid, where each grid cell corresponds to a floor tile of fixed size.

After generating the floor grid, the structural boundaries of the room are constructed by placing walls along the outer edges of the floor layout. For each wall segment, the algorithm probabilistically decides whether to place a standard wall element, a window, or a door based on predefined probabilities. To ensure scene accessibility, at least one doorway is always placed in the generated layout.

Following the wall placement, ceiling elements are instantiated above each floor tile at a fixed height, completing the basic architectural structure of the room. Furniture objects are then added in two stages. First, wall-mounted furniture elements such as shelves or cabinets may be placed along interior walls with a configurable probability. Second, free-standing furniture is placed within the interior of the room. Candidate furniture positions are sampled randomly within the room bounds while ensuring that objects do not intersect with previously placed items. This is achieved using bounding box intersection tests, which enforce collision-free placement of objects and produce physically plausible layouts.

Each random function is called from a single seed. It is possible to predefine this seed, making the whole generation deterministic. This can be useful in case the scene needs the be regenerated at a later stage.

\subsection{Asset Styles}

To enable the generation of diverse indoor environments, the procedural generation system uses configurable asset collections implemented through Unity ScriptableObjects. Each asset configuration defines a consistent visual style for a room and specifies the prefabricated models used for structural elements and furniture.

A style definition contains references to prefabs for floors, walls, ceilings, doors, and windows, as well as collections of furniture objects that can be placed either against walls or within the interior of the room. In addition, the configuration specifies parameters such as grid size, ceiling height, and the probabilities used for placing architectural elements. Furniture placement density and object categories can also be customized within the style definition.

This modular design allows different architectural styles or furniture themes to be generated simply by switching the corresponding asset configuration. As a result, the dataset can contain visually diverse scenes while using the same underlying procedural generation algorithm.

\subsection{Automatic Scan Configuration}

After a room layout has been generated, the scanning system is automatically configured to capture the scene. The scanning region is determined by computing the bounding box of the generated room. The virtual scanner is then positioned at the centre of this bounding box and configured to capture the entire scene.

This automated setup ensures that each generated room is scanned under consistent conditions without manual intervention. The scanner parameters, such as resolution and range, can be adjusted globally to emulate different real-world scanning devices. Once configured, the virtual scanning process described in Section~\ref{sec:virtual_scanner} is executed to generate the point cloud representation of the scene.

By combining procedural layout generation with automatic scanner placement, the system enables large numbers of scenes to be generated and scanned efficiently. Each generated scene contributes a new set of partial object scans, full-scene scans, and associated ground-truth data to the \textit{V-Scan} dataset.

\subsection{Dataset Metrics}

\begin{figure*}
    \centering
    \includegraphics[width=0.8\linewidth]{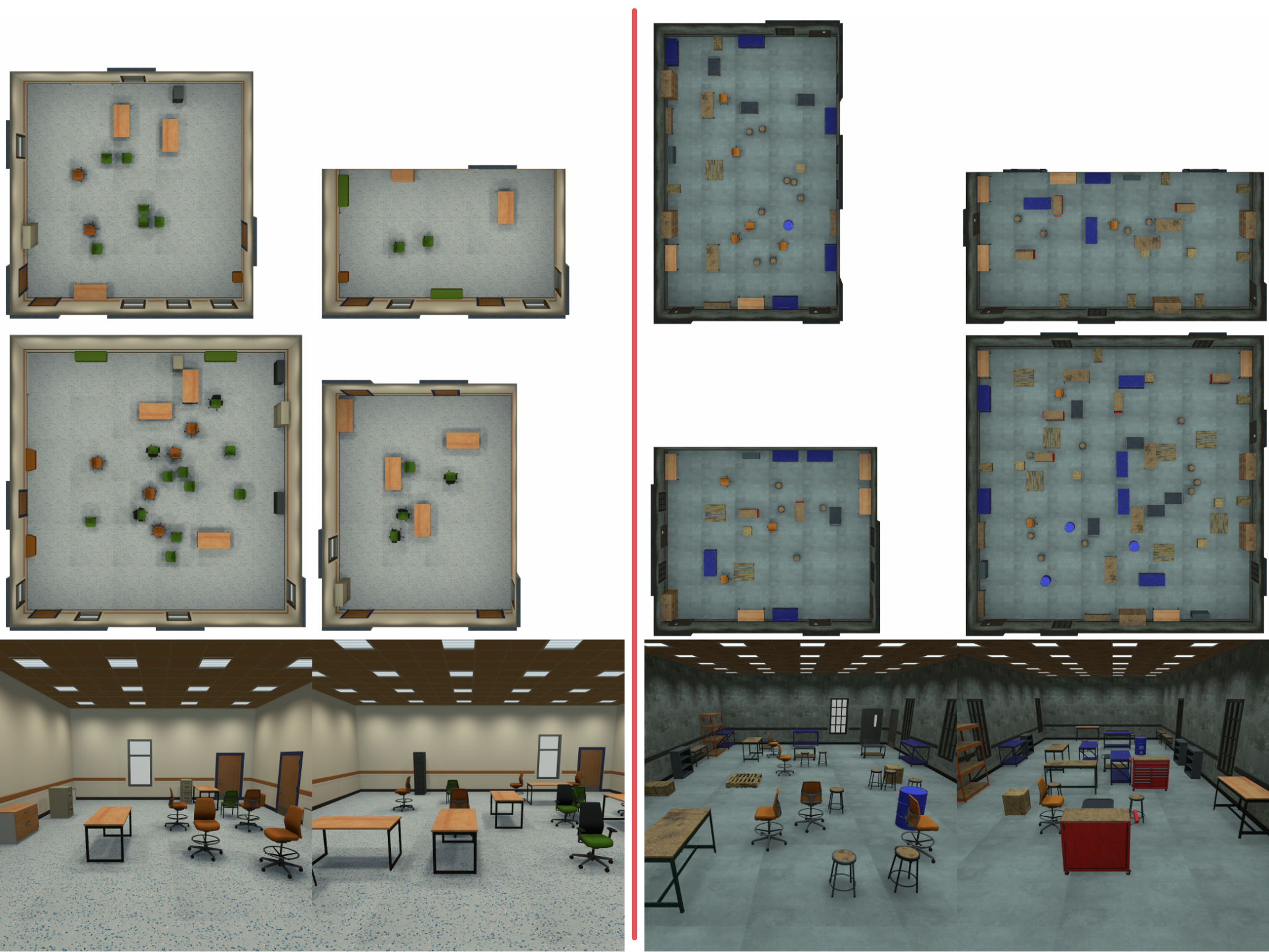}
    \caption{An overview of the different layouts, with top-down and perspective views respectively}
    \label{fig:dataset}
\end{figure*}

The dataset consists of +-100 furnished virtual scans, all including objective ground truth for both the objects and the scene. Figure \ref{fig:dataset} shows a sample of 2 different environment styles.

\section{Conclusion}

This paper presented a virtual scanning framework for generating synthetic 3D point cloud datasets of indoor environments. The proposed system simulates realistic scanning behaviour using configurable parameters derived from real-world scanning devices, enabling the generation of point clouds that include realistic measurement noise, visibility constraints, and occlusions. By integrating the scanner simulation within a procedural scene generation pipeline, the framework enables efficient creation of diverse indoor scenes that can be automatically scanned without manual intervention.

Using this system, we introduced the \textit{V-Scan} dataset, which contains synthetic indoor scans together with rich ground-truth information. The dataset includes furnished and empty scene scans, object-level partial point clouds, oriented bounding boxes, and voxel-based occlusion annotations. This combination of observed geometry and ground-truth information provides valuable supervision signals for tasks such as object completion, occlusion reasoning, and scene reconstruction.

The results demonstrate that virtual scanning can provide a scalable alternative to real-world data acquisition while maintaining realistic scanning characteristics. Future work will focus on extending the framework to support mobile scanning trajectories and increasing the diversity of procedurally generated environments. These extensions will further improve the realism and applicability of synthetic datasets for learning-based 3D perception methods.

{
	\begin{spacing}{1.17}
		\normalsize
		\bibliography{Virtual_Scanner} 
	\end{spacing}
}

\end{document}